\documentclass[conference]{IEEEtran}
\IEEEoverridecommandlockouts

\usepackage{cite}
\usepackage{amsmath,amssymb,amsfonts}
\usepackage{algorithmic}
\usepackage{graphicx}
\usepackage{textcomp}
\usepackage{xcolor}
\usepackage{booktabs}
\usepackage{enumitem}

\def\BibTeX{{\rm B\kern-.05em{\sc i\kern-.025em b}\kern-.08em
    T\kern-.1667em\lower.7ex\hbox{E}\kern-.125emX}}
\begin{document}

\title{Autonomy and Safety Assurance in the Early Development of Robotics and Autonomous Systems}

\makeatletter
\newcommand{\linebreakand}{%
\end{@IEEEauthorhalign}
\hfill\mbox{}\par
\mbox{}\hfill\begin{@IEEEauthorhalign}
}
\makeatother
	
\author{\IEEEauthorblockN{Dhaminda B. Abeywickrama*}
\IEEEauthorblockA{\textit{Department of Computer Science} \\
\textit{The University of Manchester}\\
Manchester, UK \\
dhaminda.abeywickrama@manchester.ac.uk}
\and
\IEEEauthorblockN{Michael Fisher}
\IEEEauthorblockA{\textit{Department of Computer Science} \\
\textit{The University of Manchester}\\
Manchester, UK \\
michael.fisher@manchester.ac.uk}
\linebreakand 
\IEEEauthorblockN{Frederic Wheeler}
\IEEEauthorblockA{\textit{Regulatory Support Directorate} \\
\textit{Amentum}\\
Warrington, UK \\
frederic.wheeler@global.amentum.com}
\and
\IEEEauthorblockN{Louise Dennis}
\IEEEauthorblockA{\textit{Department of Computer Science} \\
\textit{The University of Manchester}\\
Manchester, UK \\
louise.dennis@manchester.ac.uk}}

\maketitle

\section*{Executive Summary}
This report provides an overview of the workshop\footnote{This workshop was organised by Dhaminda B. Abeywickrama, Kayleigh Jackson, Michael Fisher, Harry Newton, Frederic Wheeler, and Louise Dennis.} titled ``\textit{Autonomy and Safety Assurance in the Early Development of Robotics and Autonomous Systems}'', hosted by the Centre for Robotic Autonomy in Demanding and Long-Lasting Environments (CRADLE) on September 2, 2024, at The University of Manchester, UK. 
The event brought together representatives from six regulatory and assurance bodies across diverse sectors to discuss challenges and evidence for ensuring the safety of autonomous and robotic systems, particularly autonomous inspection robots (AIR). The workshop featured six invited talks by the regulatory and assurance bodies. CRADLE aims to make assurance an integral part of engineering reliable, transparent, and trustworthy autonomous systems. Key discussions revolved around three research questions:
\begin{enumerate}
	\item \textit{Challenges in Assuring Safety for AIR:} There are several challenges, such as managing human-robot interactions effectively and the need for transparency and explainability so that users can understand AI decision-making. Ensuring testing and verification and validation (V\&V) is essential as well as the need for tailored assurance approaches. Furthermore, established benchmarks and standards are essential. 
	\item \textit{Evidence for Safety Assurance:} Heterogeneous V\&V methods, including simulations, physical testing, and real-world experiments, are needed to ensure safe operation under diverse conditions. Risk management processes and adherence to industry standards are important. Additionally, there is a need for human oversight and evidence to demonstrate that operators can effectively intervene.
	\item \textit{How Assurance Cases need to differ for Autonomous Systems:} As AIR operates in environments with dynamic, unpredictable conditions, assurance cases need to emphasise robust perception, decision-making, adaptability, and safety mechanisms that account for the absence of human oversight. There is also a need for accountability frameworks to address legal and societal trust concerns arising from failures.
\end{enumerate}

Following the invited talks, the breakout groups further discussed the research questions using case studies from ground (rail), nuclear, underwater, and drone-based AIR. 
This workshop offered a valuable opportunity for representatives from industry, academia, and regulatory bodies to discuss challenges related to \textit{assured autonomy}. 
Feedback from participants indicated a strong willingness to adopt a design-for-assurance process to ensure that robots are developed and verified to meet regulatory expectations. 
\emph{Reference assurance cases} can serve as standardised templates or examples for developing assurance cases across various industries, facilitating alignment with regulatory standards and supporting certification \cite{AbeywickramaDennis2024}. They hold the potential to more efficiently develop the assurance cases and ensure best practice is maintained. At CRADLE, we aim to build upon this foundation through the concept of reusable \textit{assurance patterns} \cite{AbeywickramaDennis2024}.

\section{Introduction}
The Cross-Sector Workshop\(^1\), titled ``\textit{Autonomy and Safety Assurance in the Early Development of  Robotics and Autonomous Systems}'', took place on 2 September 2024 at The University of Manchester, UK. Hosted by the Centre for Robotic Autonomy in Demanding and Long-Lasting Environments (CRADLE), the event brought together a diverse group of six regulatory and assurance bodies: the Health and Safety Executive (HSE), the Office for Nuclear Regulation (ONR), the Rail Safety and Standards Board (RSSB), the Maritime and Coastguard Agency (MCA), the Environment Agency (EA), and the Civil Aviation Authority (CAA).

The CRADLE Prosperity Partnership \cite{CRADLE} combines Amentum's extensive industrial experience in applied robotics and autonomous systems across sectors with the University of Manchester's unique expertise in Robotics and AI research, creating an internationally leading, sustainable, collaborative research centre.  
Within CRADLE, we aim to move assurance from being an afterthought to becoming an integral part of engineering reliable, transparent, and trustworthy autonomous systems that demonstrate the behaviours and evidence that regulators require. This workshop provided an important step in shaping the direction of CRADLE’s work, and marked the start of an ongoing series of engagements with regulators, which will continue in the coming years.

The workshop began with Michael Fisher, Academic Director of CRADLE, welcoming all invited guests and participants. James Kell and Simon Watson, the Industry and Academic Co-Directors, introduced the CRADLE programme. Louise Dennis, Academic Lead for the Assurance Work Package, provided an overview of assurance activities conducted within CRADLE, while Kayleigh Jackson, Industry Project Manager, moderated the sessions. 
The two main plenary sessions featured invited talks by six representatives: Nicholas Hall (HSE), Vincent Ganthy (RSSB), Tom Eagleton and Paolo Picca (ONR), Sam Hodder (MCA), Nicholas Bloomfield (EA), and Helen Leadbetter (CAA). Additional regulators in attendance included Mary Marshall (HSE) and Clive Tunley (ONR). After the invited talks, participants joined breakout sessions exploring a range of use case domains being investigated within CRADLE, with a focus on autonomous inspection robots (AIR) for ground, nuclear, underwater, and aerial applications. The breakout sessions were facilitated by Frederic Wheeler (Industry Lead for the Assurance Work Package), Louise Dennis, Michael Fisher, and Harry Newton (Industry Co-Lead for Assurance). 
The workshop concluded with a summary of key outcomes, and a lab tour led by Paul Baniqued, Academic Project Manager.

The rest of this report is organised as follows. Section II provides definitions used in this report and describes the four case studies utilised during the breakout sessions. Section III details the key points addressing the three main research questions, and identifies some areas of common ground across sectors. Finally, Section IV offers concluding remarks and outlines future perspectives.

\section{Background}
\subsection{Definitions}
\textit{Autonomy} is defined as ``the capacity of a system to achieve goals while operating independently from external control'' \cite{NASA2015}. 
An autonomous system is ``one that makes and executes a decision to achieve a goal without full, direct human control'' \cite{Rouff2022}. Sheridan and Verplank \cite{Sheridan1978} describe ten levels of automation, with the final level being able to operate without human supervision, which could be construed as fully autonomous. 
Topcu et al. \cite{Topcu2020} define \textit{assured autonomy} as ``understanding and
mitigating risks of operating autonomous systems in our society''. These risks can encompass the safety, security, and reliability of an autonomous system. In this workshop, our focus was primarily on \textit{safety}.

\textit{Assurance} involves providing justified confidence that a component, system, or service has the necessary assurance properties \cite{Denney2023}. Assurance activities include complying with standards, achieving certification, and performing appropriate verification and validation (V\&V)~\cite{Abeywickrama2024}. An \textit{assurance case} provides a structured argument, supported by evidence, to justify some key property (e.g., safety, security) of a system \cite{Denney2023}. 

\subsection{Case Studies}
We now describe the four case studies that were used during the breakout sessions to explore the three research questions (expanded in Section III).

\subsubsection{Ground (Rail) Autonomous Inspection Robot Use Case}
This scenario involves inspecting landslips and vegetation around railway lines. 
An unmanned ground vehicle (UGV) such as the Clearpath Husky, equipped with Light Detection and Ranging (LiDAR) and cameras, patrols a section of track to monitor any changes in the terrain. 
The system can autonomously provide early warnings of potential landslips and report on overgrown vegetation that could interfere with railway tracks or obscure geotechnical features. 
The UGV must avoid obstacles and maintain the required safety distance during navigation at all times. Additionally, it must prevent its battery charge from reaching a critical threshold. If the UGV detects any signs of landslips or vegetation growth, it must report these to the central server. 
Drone technology can additionally be employed to conduct an aerial inspection of the site and collect mapping data through an onboard sensor package. These identified locations can then be used to guide the deployment of ground inspection. 
Other inspection scenarios include the inspection of rail integrity and train wagons. For rail integrity, a robot mounted on the rail tracks (e.g., RIIS005 railway inspection robot) can perform inspections to ensure their integrity. It can track the geometry of the rails, monitor the rail profile, measure wear, assess clearances, and detect defects in fasteners or on the rail surface. For train wagons, a robot dog (e.g., ANYmal) equipped with high-resolution cameras and sensors can detect minor cracks and signs of wear on the wheel axles.

\subsubsection{Nuclear Autonomous Inspection Robot Use Case}
The mission here involves performing regular inspections in a space containing radioactive material, with the primary objective of capturing high-quality images for analysis and assessment. 
A study of risk mitigation strategies has identified an autonomous ground robot as a suitable solution, effectively balancing mission effectiveness with safety considerations \cite{Benjumea2024}. 
An AgileX Scout Mini is used for this mission, which is a commercial robot that includes a 4WD mobile chassis and a Robot Operating System (ROS) development kit. 
It is equipped with high-performance industrial control, high-precision LiDAR, and multiple sensors based on the AgileX Robot ROS ecosystem. These features enable it to perform various functions, such as mobile robot motion control, communication, navigation, and map building \cite{SCOUTMINI}. 
The system architecture integrates a traditional control system developed in ROS, together with a safety system \cite{Anderson2023}. The safety system consists of an agent responsible for monitoring compliance with functional safety properties and taking corrective actions if a violation is detected. 
Formal methods are employed to verify this agent, which allows to formally demonstrate that the system performs as expected. Mission assumptions include receiving a detailed inspection point map, positioning the charging station outside the inspection room to ensure accessibility, and maintaining stable radiation levels within the inspection room once the robot is deployed.

\subsubsection{Underwater Autonomous Inspection Robot Use Case}
This scenario involves performing integrity assessments on pier poles. Currently, divers carry out underwater inspections of pier pilings and sheet piles to evaluate their condition and estimate their remaining lifespan. These evaluations involve clearing marine growth, conducting visual inspections, capturing photographs, taking dimensional measurements, performing non-destructive evaluation (NDE) tests, and occasionally collecting samples. The pier structures may consist of materials such as wood, concrete, steel, or iron. 
Two underwater robots are employed for visual and structural inspections of the poles using grippers. Each robot is equipped with a stereo camera and lights for inspection and 3D reconstruction, a Doppler Velocity Log (DVL), an Inertial Measurement Unit (IMU), and a barometer for state estimation, along with a two-finger gripper. They also include a high-performance computer board, a battery pack providing up to four hours of autonomy, internal sensors for monitoring temperature, humidity, and current, and a communication system. 
The robots are deployed at a predetermined location, from which they autonomously navigate to inspect all the poles. Each pole is mapped, automatically visually checked, and pinched at different heights or identified points of interest to assess its integrity. If the battery is low or a fault occurs, the robot automatically returns to the surface. After the robot is retrieved, the data is transferred, and reports are generated automatically.

\subsubsection{Drone Fire Inspection Use Case}
The mission here is to patrol a designated area by flying over a set of predefined locations specified by GPS coordinates and height~\cite{Araujo2024}. The hardware system is based on the DJI Matrice 600 Pro UAV, which includes a vision-based perception system consisting of a thermal camera and a depth camera to accurately identify and localise fires~\cite{Araujo2024}. The software for the UAV is developed using ROS. The key modules of the system include the Flight Planner, which controls the UAV's navigation based on sensor inputs and directs it to follow a set of waypoints. 
The Fire Detection System identifies fires using data from thermal and depth cameras, providing this information to the Flight Planner and other modules. 
The Spray Aim Module directs the fire suppressant system by adjusting the UAV's position and activating the pump to extinguish the fire. 
The Ground Station handles mission control, which includes starting, stopping, and overriding tasks. Additionally, the Battery Monitor and Water Monitor track the UAV’s battery level and water supply, respectively, and send this information to the Flight Planner and Spray Aim modules~\cite{Araujo2024}. 
The UAV operates autonomously, executing a predefined patrol of specific GPS locations, detecting fires autonomously, navigating to them, and discharging fire suppressant without human intervention~\cite{Araujo2024}. 
It is equipped with redundant systems like IMUs and GPS to enhance flight safety. The system includes a battery monitoring component that ensures the UAV can safely return to its starting point or land immediately if critical battery thresholds are reached. Safety is further enhanced by runtime verification, where the system is continuously monitored to ensure it operates within acceptable parameters.

\section{Research Questions}
Before the workshop, we targetted three \emph{research questions}.
\begin{enumerate}
	\item \textbf{RQ1:} What challenges do you foresee in assuring the safety of an `autonomous' robot (e.g., an autonomous inspection robot)? 
	\item \textbf{RQ2:} What types of evidence do you expect to see in an assurance case supporting the safety claims of an `autonomous' robot (e.g., an autonomous inspection robot)?
	\item \textbf{RQ3:} Should an assurance case for an `autonomous' robot (e.g., an autonomous inspection robot) differ from traditional safety cases for human-operated systems? If so, how? 
\end{enumerate}
The six invited talks focused on the first two research questions, while the breakout session discussed all three.

To maintain the anonymity of the regulatory and assurance bodies, we have organised the participants’ responses to the three research questions according to the CRADLE project’s work packages --- referred to as ‘themes’ throughout this report: Components, Architectures, Interactions, Assurance, and Demonstrators \cite{CRADLE}. The \textit{Components} theme focuses on improving the reliability of subsystems to ensure mission success in resilient robotic autonomy \cite{CRADLE}.
The \textit{Architectures} theme focuses on designing resilient and verifiable software architectures to enable the reliable deployment of autonomous robots in demanding and long-lasting scenarios \cite{CRADLE}.
Meanwhile, the \textit{Interactions} theme addresses challenges in human-robot interactions, aiming to foster trustworthy and effective teamwork \cite{CRADLE}.
The \textit{Assurance} theme is dedicated to generating stronger evidence and arguments to support the ethical, safe, and secure design and implementation of robotics and autonomous systems \cite{CRADLE}.
Finally, the \textit{Demonstrators} theme showcases next-generation solutions across a range of cyber-physical environments, paving the way for the real-world deployment of robotic platforms \cite{CRADLE}.

Key insights on RQ1-RQ3 are detailed in subsections IIIA–-IIIC, while subsection IIID highlights areas of common ground for each research question across sectors.

\subsection{RQ1: Key Challenges for Assured Autonomy}

The safety assurance of AIR can present numerous challenges (see Fig.~\ref{fig:challenges}). 

\paragraph{Components} 
Ensuring that off-the-shelf components meet safety standards is challenging, particularly because of the variability and uncertainty inherent in operational environments. The completeness and correctness of the operational domain and its boundaries must be ensured, with adequate coverage to demonstrate safety throughout the robot's lifecycle.
\begin{figure}[!t]
	\centering
	\includegraphics[width = 1.03\columnwidth]{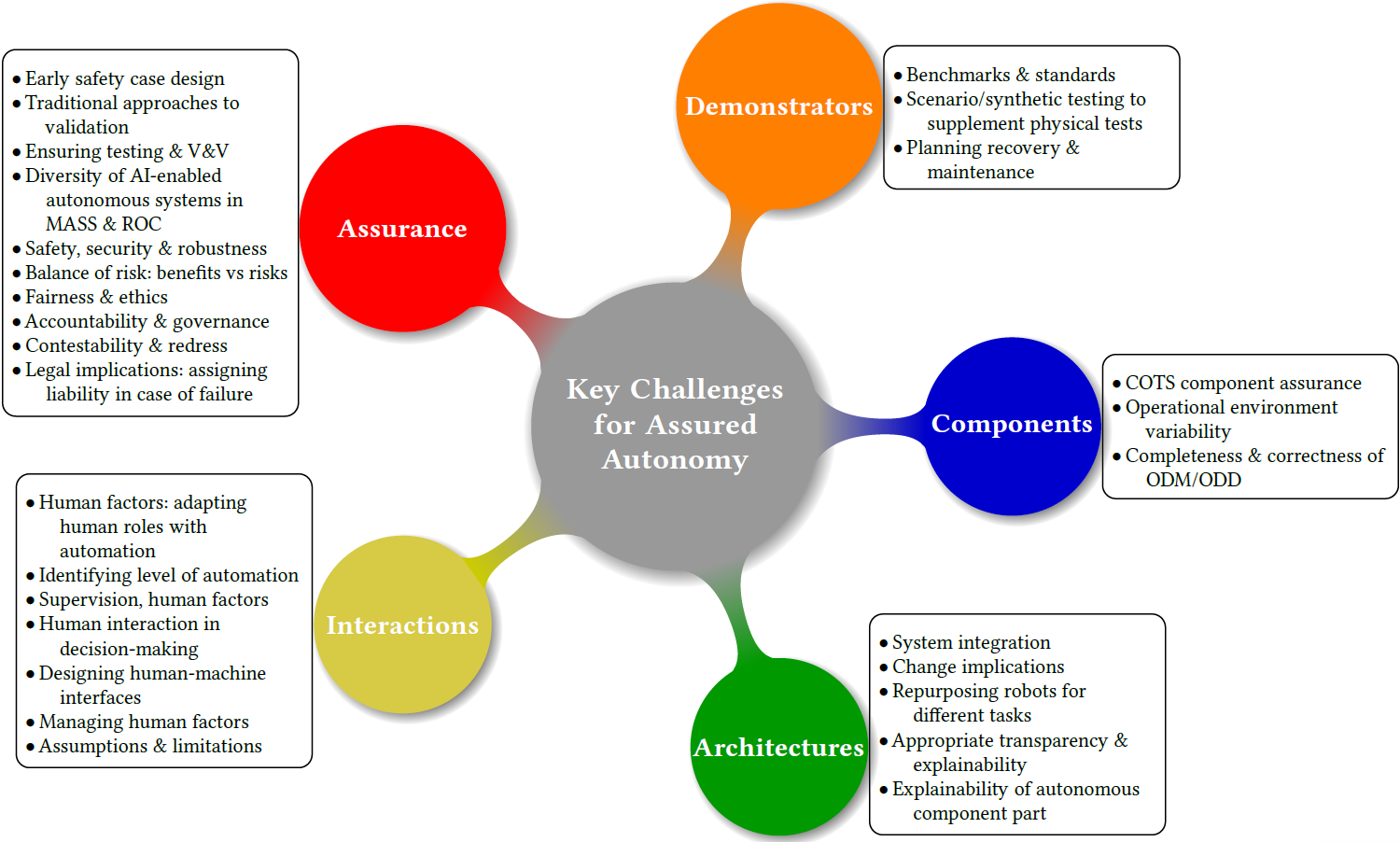}
	\caption[Key challenges for assured autonomy.]{Key challenges for assured autonomy. \footnotemark}
	\label{fig:challenges}
\end{figure}

\paragraph{Architectures}
Integrating the robot both in terms of subsystems and the overall system is a key challenge (see Fig.~\ref{fig:challenges}).
Careful attention must be given to addressing the implications of introducing new or changed functions.
Difficulties arise when repurposing robots for tasks beyond their original safety limits.
Transparency and explainability need to be ensured so that users can understand AI decision-making and the role of the core autonomous component.

\paragraph{Interactions} 
Interactions between humans and robots introduce another layer of challenges. 
Understanding the changing roles of humans in systems with increased automation presents a key challenge.
Identifying the level of automation for each system function and determining the required training is significant.
The role of human supervision and the challenges of human-robot interaction must be addressed.
Designing human-machine interfaces that provide operators with sufficient information is important.
Furthermore, managing human factors, such as reduced situational awareness, and clearly communicating the assumptions and limitations are essential to effective human-robot collaboration and safety (Fig.~\ref{fig:challenges}).

\paragraph{Assurance}
Focusing on the safety case at the early stages of design as an enabler is important. 
Issues with traditional validation methods in autonomous systems require careful consideration.
Ensuring testing and V\&V is essential to confirm the robot’s functionality is robust.
The diverse nature of AI systems necessitates tailored assurance approaches, with attention to cybersecurity during both the design and development phases.
Ensuring safety, security, and robustness under adverse conditions is vital, alongside weighing the benefits of autonomous robots against the new risks they introduce. 
Addressing fairness and ethics is vital to prevent unfair treatment, while establishing accountability and governance --- including making operators responsible for AI use --- is necessary for ethical operation.
Enabling contestability and redress for users to challenge harmful or risky AI decisions is crucial, as is identifying liability when autonomous systems fail.
Furthermore, evaluating the safety of decisions made by the robot is necessary to maintain trust (see Fig.~\ref{fig:challenges}). 
\footnotetext{Key: ODM/ODD - Operational Domain Model/Operational Design Domain, ROC - Remotely Operated Craft, MASS - Maritime Autonomous Surface Ships.}

\paragraph{Demonstrators} 
The availability and appropriateness of established benchmarks and standards are significant. Scenario-based and synthetic testing is needed to supplement physical testing, and maintenance and recovery plans for robot failures must also be developed to ensure operational continuity.

\subsection{RQ2: Types of Evidence for Assured Autonomy}
To build a robust assurance case supporting the safety claims of an AIR, various types of evidence are essential (see Fig.~\ref{fig:evidence}).
\begin{figure}[!t]
	\centering
	\includegraphics[width = 1.1\columnwidth]{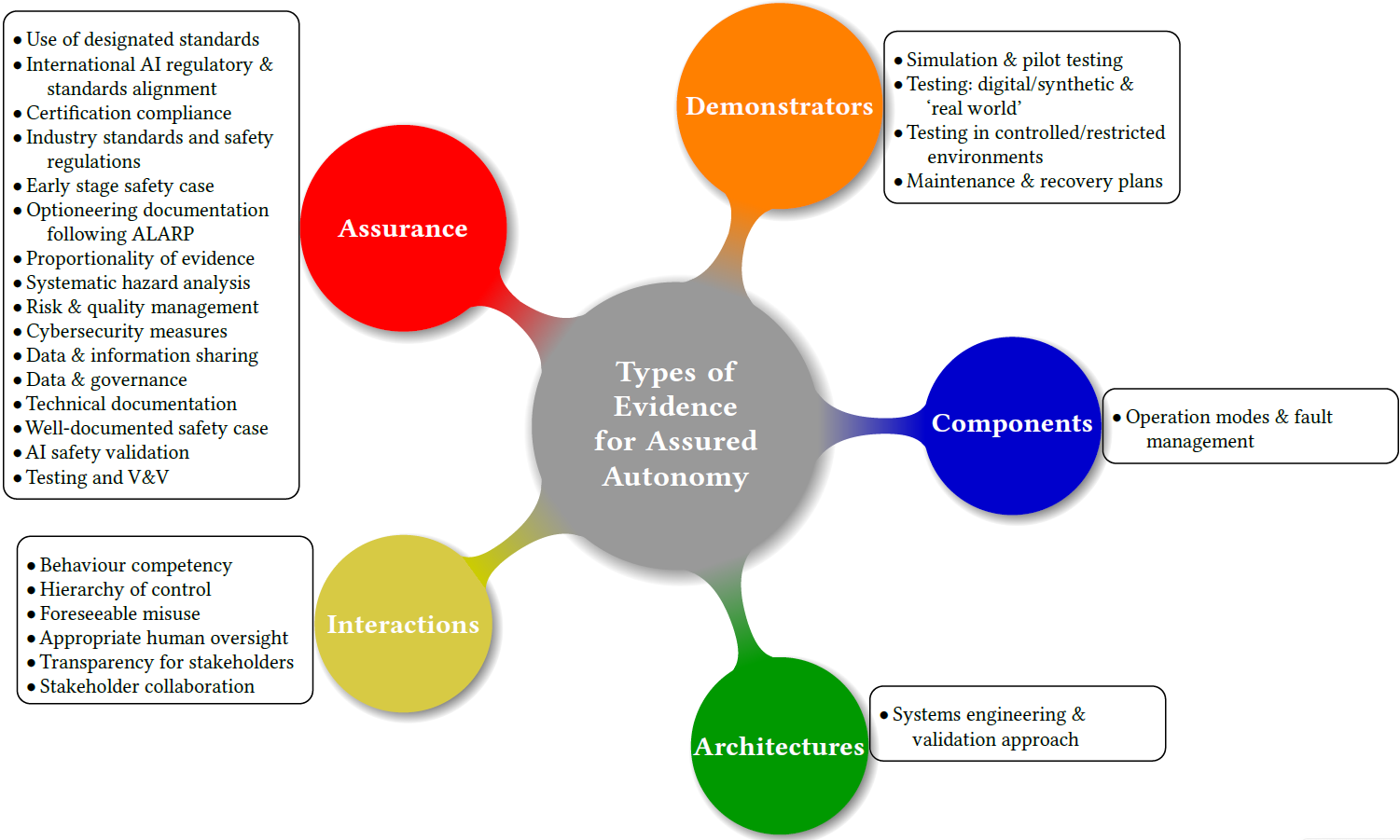}
	\caption{Types of evidence for assured autonomy.}
	\label{fig:evidence}
\end{figure}

\paragraph{Components} The assurance case must consider all modes of operation, including both normal and fallback (failure) modes. 

\paragraph{Architectures} The assurance case must highlight the systems engineering and validation approach. It should also be explicit about the \emph{assumptions} inherent within it --- assumptions about the system's capabilities and assumptions about the system's environment.

\paragraph{Interactions} 
Interaction-related evidence should include proof of the behavioural competency of the autonomous system; the application of hierarchy of control for autonomous equipment; measures addressing foreseeable operator misuse and robot misbehaviour; and guarantees that human operators can intervene when needed to provide oversight (see Fig.~\ref{fig:evidence}).
Documentation should describe the robot's decision-making processes to stakeholders, supplemented by evidence of collaboration with stakeholders and effective interface management.

\paragraph{Assurance} 
Designated standards need to be adhered to where possible to align with existing safety benchmarks.
Additionally, where possible there must be alignment with international regulatory frameworks and standards for AI; conformance with relevant certification specifications; and compliance with industry standards, health and safety regulations. 
An early-stage safety case, which includes evidence from the early stages of design, is vital. This evidence should encompass systematic hazard analysis and system security process analysis to understand potential risks. It should also include optioneering, ensuring that multiple design solutions are considered in accordance with the As Low As Reasonably Practicable (ALARP) principle. Furthermore, the proportionality of evidence must align with the complexity of the robot's function. 
The safety case must be well-documented, demonstrating a risk-based approach.
Risk and quality management—proof of processes in place to identify and mitigate risks—is significant, alongside cybersecurity measures integrated during the design and development of the autonomous system.
Evidence of proper data handling, information sharing, and governance, as well as technical documentation describing system construction, testing, and validation, is essential. 
Safety validation testing must demonstrate the independence of the AI and safety systems, and testing and V\&V must demonstrate robust RAS functionality under varied operational conditions.

\paragraph{Demonstrators} 
Evidence needs to include simulation and pilot testing for data generation and digital assurance. Testing should incorporate both digital/synthetic and physical/real-world testing, and it should be performed in both controlled/restricted environments before wider deployment. Additionally, there needs to be evidence of maintenance, testing, recovery from malfunctions, and equipment reliability in harsh environments.

\subsection{RQ3: Assurance Cases for Autonomous Robots}
An assurance case for an AIR must address the unique challenges posed by \textit{autonomy}. The breakout groups examined these challenges through four case studies described in subsection IIB.

\paragraph{Ground/Rail Use Case} Autonomous robot behaviour must be explicitly addressed to ensure the robot does not stray into operational areas, particularly given the heightened risks posed by new technologies, such as theft. Unlike traditional systems, autonomous robots that utilise Commercial Off-The-Shelf (COTS) equipment require additional responsibility and integration. Autonomous robots must account for external, uncontrolled factors, such as trespassers, which human-operated systems can often manage more effectively. 
The lack of established standards or benchmarks for autonomous robots complicates regulatory approval, presenting challenges. 
These robots must also incorporate high-integrity safety systems, such as LiDAR, to ensure collision avoidance. They need to handle unexpected events and high-consequence scenarios, ensuring decisions are both legally defensible and trustworthy.

\paragraph{Nuclear Use Case}The assurance case should limit safety claims to the specific parts of the system that are autonomous, integrating proven technologies to enhance reliability. Autonomy introduces risks and new hazards that differ from those of human-operated systems, necessitating proportional safety claims supported by solid evidence. Accurate data processing is critical, as AI and Simultaneous Localisation and Mapping (SLAM) errors could lead to missed inspections. 
The assurance case must specify how the system handles failures autonomously without human intervention. Additionally, the assurance case must consider hardware measures to compensate for software errors, as well as address risks associated with battery disposal and long-term degradation (security).

\paragraph{Underwater Use Case} The assurance case must emphasise the robot's ability to perceive, interpret, and safely interact with unpredictable environments without human oversight. It must introduce new considerations for accountability, especially in cases of failure, and ensure the accuracy, robustness, and reliability of sensor data and AI decision-making. 
Varying environmental conditions, such as offshore versus inshore operations, must also be factored in, demonstrating the robot’s ability to adapt autonomously to different challenges. 
Furthermore, the assurance case must demonstrate the system's ability to consistently perform in repetitive tasks without degradation in performance or increased risk. It also needs to address the dual responsibility of keeping humans safe and preventing the robot from inadvertently causing harm.

\paragraph{Drones Use Case} The assurance case must account for the ways in which autonomy reduces human error but introduces concerns about machine decision-making in uncertain or unstructured environments. It must address both technical failures and unsafe behaviours stemming from autonomous decisions, requiring robust algorithms and monitoring systems. Ensuring safe interactions between humans and the autonomous system is critical, as is addressing cybersecurity risks associated with data-driven AI. 

\subsection{Discussion: Cross-Sector Common Ground}
Subsections IIIA--IIIC provided insights into the three research questions explored during the workshop. In this subsection, we aim to identify common ground for each research question across sectors.

The participants highlighted several common challenges in ensuring the safety of AIR (RQ1). Managing human-robot interactions, such as defining the appropriate level of human supervision and ensuring clear communication, emerged as a recurring concern across several sectors. 
Another widely acknowledged challenge was ensuring the safety, security, and reliability of autonomous systems through robust testing, verification, and developing maintenance and recovery plans. The need for transparency and explainability of AI decision-making was broadly emphasised, as these qualities are essential for fostering trust and accountability. Furthermore, establishing safety standards and ensuring regulatory compliance, including benchmarks and tailored assurance approaches, were identified as significant challenges.

Participants also identified shared expectations for the types of evidence (RQ2) necessary for assured autonomy in AIR. 
Heterogenous V\&V methods were emphasised, including simulations, physical testing, and real-world experiments to ensure safe operation under diverse conditions. Risk management processes (e.g., hazard analysis) were identified as essential for mitigating safety concerns. 
Evidence of compliance with industry standards and regulatory requirements was widely noted, ensuring alignment with established benchmarks. Additionally, human oversight was a recurring theme, with evidence required to demonstrate that operators can effectively intervene and that safeguards are in place to prevent misuse or address malfunctions. 

Several common strategies were identified regarding how assurance cases (RQ3) for autonomous robots should differ from traditional safety cases for human-operated systems. 
A key concern highlighted by several bodies was that AIR operates in environments with dynamic, unpredictable conditions, requiring assurance cases to emphasise robust perception, decision-making, and adaptability. 
Unlike human-operated systems, autonomous robots must respond to environmental variations and unforeseen events, with safety mechanisms accounting for the absence of human oversight. 
Assurance cases need to integrate accountability frameworks that address legal and societal trust issues arising from failures.
A shared concern was ensuring the reliability of AI and sensor systems, which need to be proved to perform consistently and accurately. Assurance cases also need to highlight the robot’s ability to autonomously manage failure scenarios, with evidence of robust mechanisms such as hardware redundancies that compensate for software faults. Although sector-specific nuances were noted---for example, collision avoidance for ground robots, cybersecurity concerns for drones, and battery degradation in nuclear robots---these commonalities provide a foundation for a unified approach to assurance.  

\section{Conclusion and Outlook}
\begin{sloppypar}
	This workshop offered a valuable opportunity for representatives from industry, academia, and regulatory bodies to come together and discuss challenges of \textit{assured autonomy}. 
	Feedback from participants indicated a strong willingness to adopt a design-for-assurance process to ensure that robots are developed and verified to meet regulatory expectations. 
\end{sloppypar}

As discussed, assuring AIR presents significant challenges due to their inherent complexity, the unpredictability of emergent behaviours, and the uncertainties inherent in their operating environment. Traditional methods of V\&V are often inadequate for systems that learn and adapt, underscoring the need for innovative approaches. Concurrently, regulatory frameworks face increasing pressure to evolve and address these unique challenges. 
In this context, structured frameworks or templates can offer a practical approach to standardising the development of assurance cases across industries. 

A \textit{reference assurance case} serves as a template or exemplar, encompassing a repository of accepted practices, logical arguments, and evidential standards \cite{AbeywickramaDennis2024}. These templates can be tailored to specific projects or systems, providing multiple benefits: reducing the time required to develop new assurance cases, improving quality through broad coverage of critical areas, and streamlining regulatory approvals \cite{AbeywickramaDennis2024}. 
Furthermore, reference cases can bridge the gap between innovative capabilities and evolving regulatory requirements. 
They could foster alignment among key stakeholders—engineers and regulators. By integrating emerging good practices \cite{RAE2019}, such as modular system designs, these frameworks could establish a baseline for navigating the complexities of assuring AIR. 

At CRADLE, we aim to build upon this foundation through the concept of reusable \textit{assurance patterns} \cite{AbeywickramaDennis2024}. Inspired by design patterns, these assurance patterns enable the reuse of safety argument structures. By adopting a \textit{corroborative assurance approach}~\cite{Webster2020} and leveraging mission specification patterns, our research seeks to address the complex V\&V requirements of instantiated assurance cases for AIR, taking a step toward addressing the challenges posed by \textit{autonomy} \cite{AbeywickramaDennis2024}. 

\section*{\small ACKNOWLEDGEMENT}
The authors sincerely thank all regulatory and assurance bodies, as well as the CRADLE members, for their participation in the workshop. Marti Morta-Garriga and Work Package 1 of CRADLE contributed to a scenario for discussions on underwater inspection, while Diana Benjumea Hernandez contributed to a scenario on nuclear inspection. Christopher Bishop provided insights into a scenario on ground (rail) inspection. 
This work is supported by CRADLE under the EPSRC grant EP/X02489X/1.

\bibliographystyle{IEEEtran}
\bibliography{Mybib}
\end{document}